\documentclass[journal]{IEEEtran}

\usepackage{comment}
\usepackage{mathtools}
\usepackage{amsmath}
\usepackage{graphicx}
\usepackage{epstopdf}

\usepackage{hyperref} 
\usepackage{cite}
\usepackage{amsmath}
\usepackage{color}

\usepackage{graphicx}
\usepackage{mathptmx}      
\usepackage{latexsym}
\usepackage[ruled,vlined]{algorithm2e}

\usepackage[misc]{ifsym}

\begin{document}

\title{A Brain-Inspired Compact Cognitive Mapping System}

\author{Taiping~Zeng,
and~Bailu~Si~\IEEEmembership{Member,~IEEE}
\thanks{T. Zeng is with Institute of Science and Technology for Brain-Inspired  Intelligence, Fudan University, Shanghai, China and Key Laboratory of Computational Neuroscience and Brain-Inspired Intelligence (Fudan University), Ministry of Education, China. (e-mail:zengtaiping.ac@gmail.com).}
\thanks{B. Si is with School of Systems Science, Beijing Normal University, 100875, China (e-mail:bailusi@bnu.edu.cn).}
\thanks{Correspondence should be addressed to Bailu Si (bailusi@bnu.edu.cn).}
}


\maketitle

\begin{abstract}
As the robot explores the environment, the map grows over time in the simultaneous localization and mapping (SLAM) system, especially for the large scale environment. The ever-growing map prevents long-term mapping.
In this paper, we developed a compact cognitive mapping approach inspired by neurobiological experiments.
Inspired from neighborhood cells, neighborhood fields determined by movement information, i.e. translation and rotation, are proposed to describe one of distinct segments of the explored environment. The vertices and edges with movement information below the threshold of the neighborhood fields are avoided adding to the cognitive map. The optimization of cognitive map is formulated as a robust non-linear least squares problem, which can be efficiently solved by the fast open linear solvers as a general problem. According to the cognitive decision-making of familiar environments, loop closure edges are clustered depending on time intervals, and then parallel computing is applied to perform batch global optimization of the cognitive map for ensuring the efficiency of computation and real-time performance. After the loop closure process, scene integration is performed, in which revisited vertices are removed subsequently to further reduce the size of the cognitive map.
A monocular visual SLAM system is developed to test our approach in a rat-like maze environment.
Our results suggest that the method largely restricts the growth of the size of the cognitive map over time, and meanwhile, the compact cognitive map correctly represents the overall layer of the environment as the standard one. Experiments demonstrate that our method is very suited for compact cognitive mapping to support long-term robot mapping. Our approach is simple, but pragmatic and efficient for achieving the compact cognitive map. 
\end{abstract}


%
\IEEEpeerreviewmaketitle

\section{Introduction}
\label{intro}

\IEEEPARstart{S}{patial} cognition provides mammals with impressive long-term navigation capabilities, which endows them to travel long distances to search foods within the wide range of environments and then unerringly return to their nests even after a long period of time. 
It is believed that mammals can learn spatial information from the surrounding environment to form an internal map-like representation in the brain, namely cognitive map, to help mammals navigate in the complex environments~\cite{tolman_cognitive_1948}.
However, even though mammals explore the same environments for tens of thousands of times and travel trajectories with thousands of miles, they always maintain their precise navigation abilities to perform various tasks, due to the efficient neural coding of the internal cognitive map in the hippocampus. 
The mechanism of mammalian spatial cognition shows a great potentiality to inspire novel algorithms to help improving the navigation ability of mobile robots. 
In this study, following our previous models~\cite{zeng2017cognitive}, which represent the environment with grid cells, and head direction cells from spatial view cells, now, we focus on how the cognitive map efficiently encodes the surrounding environments and sparsely store information during exploration process when a mobile robot builds the map of the explored environment.

In the field of robotics, mapping the environment is desirable for many applications for autonomous mobile robots, including transportation, service, delivery, and search and rescue.
Robots can effectively travel in complex environments by relying on approaches to achieve an appropriate model of the environments. While, in most of the existing graph-based algorithms, the complexity of maps grows with the length of the robot's trajectory~\cite{kretzschmar2012information}. As new vertices are constantly adding to the map, requirements of computational time and memory footprint grow over time, preventing the long-term mapping applications. So, approaches to control the size of the map, i.e. compact map, during continuous exploration of new places and the increasing time of operation, are key to unbound practical robotic applications~\cite{cadena_past_2016}.

According to recent neurobiological discoveries, the precise mechanism for mammals navigation includes place cells~\cite{okeefe_hippocampus_1971}, grid cells~\cite{hafting_microstructure_2005}, head direction cells~\cite{taube_head-direction_1990}, speed cells~\cite{kropff_speed_2015}, boundary cells~\cite{lever_boundary_2009}, etc, which increase and decrease in electrical activity to encode various information of the environment~\cite{mcnaughton_path_2006,moser_place_2008,moser_place_2015}.
However, very detailed representations of daily life environment are not used. Instead, more course information contributes to mammalian spatial navigation. Left at the supermarket, down the road, and right at the convenience store, called topographical orientation, which is adopted in the cognitive map. Neurons, called neighborhood cells, code navigation behavior at larger scales, which are thought to help the brain to differentiate distinct segments of the environment~\cite{bos2017perirhinal}.

Based on the current neurobiological experiment, we make the following hypotheses about how to store the map in the brain during long-term travel in the environment. For the first hypothesis, place cells, grid cells, and head direction cells may represent all information from spatial view cells for different views. However, just some fundamental views with important information are memorized. Go straight ahead, turn left at a crossroad, and go straight further you will see the metro station. There is only the crossroad needed to be fully remembered in this navigation process. Second, mammals rethink where they are, which depends on whether they revisit many consecutive familiar places in their memory. This is often that they perform the cognitive decision-making of familiar environments. Third, after the revisiting decision has been made, there is no need extra neural coding to represent the familiar environment. The familiar scenes are integrated without adding new vertices.

In this paper, we develop a pragmatic compact cognitive mapping solution to control the growth of the size of cognitive map, compared with stand approaches. Neighborhood cells are introduced to segment distinct parts of the explored environment with topographical orientation. Neighborhood fields are defined to describe a segment of the explored environment. In the cognitive map, the size of neighborhood fields is only decided by the movement information, i.e. translation and rotation, which is used to determine whether adding a new vertex to the cognitive map or not. This neighborhood fields control the sparsity of the cognitive map. We also formulate the optimization of the cognitive map as a constrained robust non-linear least squares problem. During the loop closure process, according to the cognitive decision-making of familiar environments, loop closure edges are clustered through time intervals. And then, we perform global parallel batch optimization using Ceres solvers and OpenMP to improve computational efficiency. Whereas, for adding sequential edges (not a loop closure edge), optimization is not performed. After optimization, scene integration is performed, in which the revisited vertices are removed to avoid encoding similar information in the cognitive map and maintain a compact map. We finally removing short edges, caused by the motion noise and measuring error.

The contribution of this paper is four-fold. 
First, inspired by neighborhood cells, we introduce a concept of neighborhood fields to segment the explored environment. The neighborhood fields are computed by the movement information and determine whether adding new vertices to the cognitive map. Since the neighborhood field is used to sparsify new adding vertices and edges, we do not need to partition the environment to reduce vertices of map, and keep the important information of cognitive representation and fidelity of representation of environment. This pragmatic approach allows us to gently trade off accuracy for computational cost including computational time and memory footprint and achieves a compact cognitive map. 
Second, following previous works, the global optimization of cognitive map is described as a robust non-linear least squares problem. Even for the large scale environment, the global optimization problem can be quickly solved by using the open fast linear solvers as a general method, not by many time iterations. 
Third, according to the cognitive decision-making of familiar environments, loop closure edges are clustered for computational efficiency. Batch optimization using Ceres solver reduces the stress of CPU load and only performs when finding a new cluster of loop closure edges. Furthermore, parallel computing with OpenMP ensures computational speed. When the robot revisits familiar image views, redundant vertices and edges are removed to ensure no extra neural coding is needed to represent the familiar environment. 
Four, a monocular visual SLAM system is implemented to evaluate the proposed compact cognitive mapping solution. We demonstrate the mapping performance of this improved monocular visual SLAM system on an iRat rodent-sized robot platform in a rat-like maze (iRat 2011 Australia dataset). Experimental results show that the size of the map is largely restricted by our approach over time. Our experiments demonstrate that our method is very suited for compact cognitive mapping to support long-term robot mapping. 


\section{Related Work}

In the context of the SLAM problem, many effective approaches have been proposed to solve robot mapping. Lu and Milios~\cite{lu1997globally} first introduced global map optimization using pose graph. The graph-based approach models the poses of the robot as vertices, and spatial constraints between poses as edges in a graph. For this standard graph-based approach, as exploring new areas and increasing operation time, the size of the map expands infinitely. Subsequently, the requirements of memory and computational capability increase boundlessly. In the worst case, the quadratic growth of memory usage with the number of variables is caused by direct linear solvers.
The following researchers are mainly focusing on improve efficiency for this solution. The sparsity structure of the matrix in the normal equations is used to enable the fast linear online solvers. Many SLAM libraries, such as g2o~\cite{kummerle_g_2011}, GTSAM, Ceres~\cite{agarwal2012ceres}, are available to solve this problem with tens of thousands of variables in just a few seconds. However, even using iterative linear solvers, the memory consumption grows linearly with the numbers of variables. Revisiting the same place many times makes this situation more complicated. As more vertices and edges continuously add to the same spatial area, this approach becomes less efficient. 
For now, there are few works to solve the question of how to store the map during long-term exploration~\cite{cadena_past_2016}. Therefore, it is desirable to achieve a long-term mapping solution that can control, or at least reduce the growth of the size of the map.

One of the most important ways to reduce the complexity of the map is vertex and edge sparsification, which trades off the accuracy of the map for memory and computational efficiency. Information-based compact pose SLAM algorithm~\cite{ila2010information} was proposed to use an information-theoretic approach to reject redundant vertices and add informative measurements to the map. An information-based criterion~\cite{kretzschmar2012information} is introduced to determine which laser scan should be marginalized in pose global optimization, which retains the sparsity for laser-based 2D pose graphs. The generic linear constraint factors~\cite{carlevaris2013generic} and the nonlinear graph sparsification~\cite{mazuran2016nonlinear} are proposed to achieve a sparse blanket based on the Markov blanket of a marginalized vertex. 

Moreover, besides marginalization to reduce the complexity of the map, other different methods to long-term operation achieve effective solution by avoiding adding redundant vertices and edges before global optimization~\cite{johannsson2013temporally}, which focuses on addressing temporal scalability of standard pose graph. This work is demonstrated on an online binocular visual SLAM system, which still focuses on indoor building-scale environments.  
A biologically inspired monocular visual SLAM, called RatSLAM, reduces experience map by removing experiences to maintain a one experience per grid square density by partitioning the environment~\cite{milford2010persistent}. When revisiting a familiar view, the current location of the robot is set to the vertex which corresponds to the familiar view, which also avoids adding duplicate experience vertices to the map. For adding every new loop closure edge, the experience map would be optimized dozens of times by iterative map relaxation for all variables.

\section{Method}
\label{Method}
In this section, an approach to achieve compact cognitive mapping is proposed to control the size of the cognitive map over the exploration time. First, global optimization of the cognitive map is formulated as a solution of non-linear least squares problem. A sparse solver is applied to solve the normal equations with high-performance Ceres solver~\cite{agarwal2012ceres}. Second, when the robot meets novel image views, inspired by neighborhood cells, neighborhood fields are used to sparsify new adding vertices and edges. Third, according to the cognitive decision-making of familiar environments, the time intervals between loop closure edges are applied to cluster loop closures edges, and further the batch global optimization is carried out. After non-linear least squares optimization, very short edges with distance threshold between two vertices are removed to filter noise from movement and measurement. Finally, when the robot meets familiar image views, considering redundant vertices between each pair of loop closure vertices, if each pair of vertices converges to the same location, revisited loop closure vertices would be removed. The familiar scenes are integrated.

\subsection{Formulation of Cognitive Map Optimization}
The optimization approach for the cognitive map is formulated as a constrained robust non-linear least squares problem. Edges are used to model spatial constraints between vertices. Sequential edges arise from odometry measurements. And loop closure edges arise from visual template matching. Additional data can be easily considered by adding new residuals. Once every loop closure cluster, we use Ceres~\cite{agarwal2012ceres} to compute a solution to
\begin{equation}
\begin{split}
\displaystyle
\min_{\mathbf{e}} &\quad \frac{1}{2}\sum_{i,j} \rho_i\left(\left\|f_i\left(e_{i},e_{j},e_{ij}\right)\right\|^2\right),
\end{split}
\end{equation}
where, $\mathbf{e}$ including all vertices $e_i$ and $e_j$ is optimized given edges $e_{ij}$, $e_i = (x_i,y_i,\theta_i)$ and $e_j = (x_j,y_j,\theta_j)$ are the vertex state. $e_{ij} = (x_{ij},y_{ij},\theta_{ij})$ describes the constraint from $e_i$ to $e_j$.  $\rho_i\left(\left\|f_i\left(e_{i},e_{j},e_{ij}\right)\right\|^2\right)$ is a residual block, where $f_i(\cdot)$ is a cost function. $\rho_i$ is a loss function, for example Huber Loss, which can be applied to decrease the influence of outliers on the global optimization of non-linear least squares problems. More specifically, cost function $f_i(\cdot)$ for a pair of vertices is computed by
\begin{gather}
\begin{split}
f_i\left(e_{i},e_{j},e_{ij}\right) 
&= \begin{bmatrix} 
e_j - e_i - e_{ij} 
\end{bmatrix} 
= \begin{bmatrix} 
x_j - x_i - x_{ij} \\ 
y_j - y_i - y_{ij} \\ 
\theta_j - \theta_i - \theta_{ij} 
\end{bmatrix}\\
&= \begin{bmatrix}
x_j - x_i - d_{ij} \cdot \cos(\theta_i + \text{heading\_rad}) \\ 
y_j - y_i - d_{ij} \cdot \sin(\theta_i + \text{heading\_rad}) \\ 
\theta_j - \theta_i - \text{facing\_rad}
\end{bmatrix}, \\
\text{s.t.} &\quad -\pi \le \theta_i < \pi ,\\
&\quad -\pi \le \theta_j < \pi,
\end{split}
\end{gather}
where $e_{ij}$ describes distance between $e_i$ and $e_j$, heading radians $\textit{heading\_rad}$, and facing radians $\textit{facing\_rad}$, $d_{ij}$ is the distance between $e_i$ and $e_j$. Since there exists relative angle radians when visual template matching, heading radians and facing radians are not the same value~\cite{ball_openratslam:_2013}. Values of $\theta_i$ and $\theta_j$ are limited to a certain range, which belongs to $[-\pi,\pi)$.

\subsection{Adding Sparse vertices through Neighborhood Fields}
Since mammals do not represent environments very detailedly, but topographical orientation, neighborhood cells~\cite{bos2017perirhinal} are modeled to differentiate distinct segments of the environment. The firing fields of neighborhood cells, called neighborhood fields, are introduced to describe one of the distinct segments of the environment. Although in the real situation, a distinct segment for mammals may be influenced by visual marks, odor, sound, movement, etc, here, in the cognitive map, we only consider movement information to determine whether the neighborhood field is large enough to form a distinct segment. The movement information of mammals includes translation $d$ and rotation $\theta$.
The neighborhood field can be defined by 
\begin{equation}
g(d,\theta) = (1 + \alpha \cdot d)(1 + \beta \cdot \theta),
\end{equation}
where, $\alpha$ is the weight for translation $d$, and $\beta$ is the weight for rotation $\theta$. If new movement information $g(d,\theta)$ is novel enough to create a new neighborhood field, i.e. $g(d,\theta) > \delta$, where $\delta$ is the threshold, a new vertex would be added to the cognitive map. For further explanation, translation $d$ and rotation $\theta$ jointly decide to whether adding new information to the cognitive map or not. If the translation $d$ is large enough to provide novel information, a new vertex is also added to the cognitive map, regardless of rotation $\theta$. When mammals make a turn, more new views should be remembered, even if the translation is small.
  
\begin{figure}[!ht]
\includegraphics[width=3.3in]{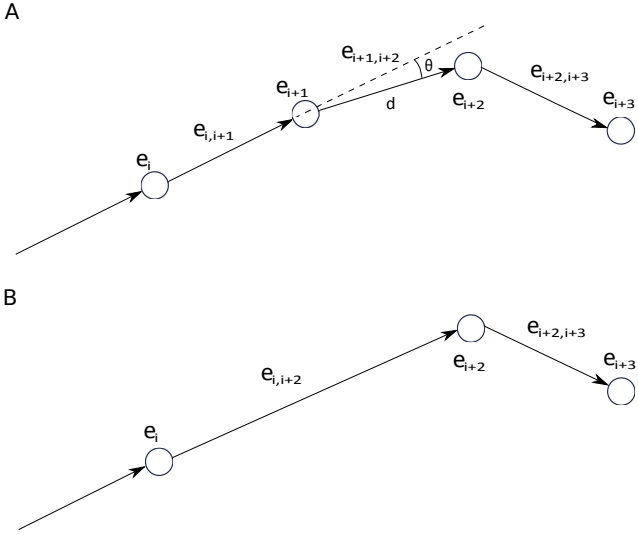}
\caption{Adding sparse sequential vertices and edges. (A) shows standard cognitive map when adding new sequential vertices and edges; (B) shows compact cognitive map.}
\label{fig_add_seq_vertices}
\end{figure}

\begin{figure}[!ht]
\includegraphics[width=3.3in]{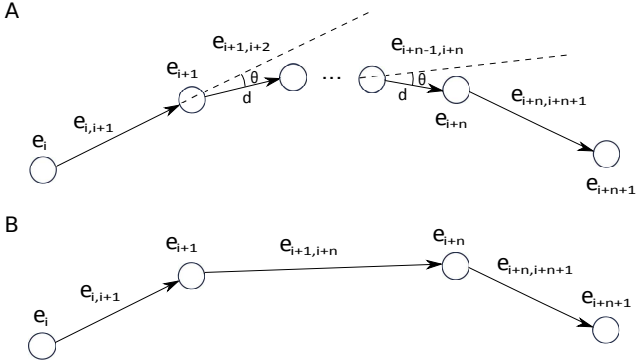}
\caption{Adding a number of consecutive sequential vertices. (A) shows standard cognitive map; (B) shows compact cognitive map.}
\label{fig_add_n_seq_vertices}
\end{figure}

We further detailedly illustrate the sparse technique using the concept of the neighborhood field. Compared with the standard approach, if the neighborhood field $g(d,\theta)$ does not meet the threshold to create a new vertex, vertex $e_{i+1}$ is removed, and edges $e_{i,i+1}$ and $e_{i+1,i+2}$ are merged into $e_{i,i+2}$, shown in Fig.\ref{fig_add_seq_vertices}. If many consecutive vertices meet threshold requirements, we would remove these consecutive vertices and merge all these edges into one, which is shown in Fig.\ref{fig_add_n_seq_vertices}.

\subsection{Clustering Loop Closure Edges}
\begin{figure}[!ht]
\centering
\includegraphics[width=3.2in]{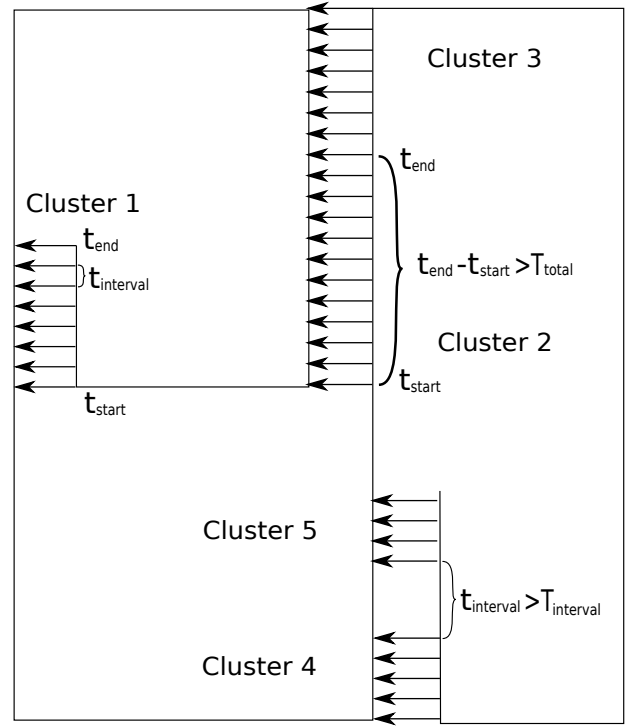}
\caption{Clustering loop closure edges. According to thresholds of time interval and total time, loop closure edges are clustered.}
\label{fig_clustering_loop_edges}
\end{figure}

According to the cognitive decision-making process, whether mammals locate in the previous visited environments are decided by consecutive familiar scene views in their memory, loop closure edges are clustered. 
An incremental way is used to group loop closure edges into a cluster based on timestamps. After clustering loop closure edges, batch optimization can be performed and there is no need to perform optimization for adding every loop closure edge. When the first loop closure edge is created, the first cluster is initialized and the start time ($t_{start}$) of the current cluster is the timestamp of the first loop closure edge. If the time interval between the current loop closure edge ($t_{end}$) and the previous loop closure edge ($t_{end-1}$) is smaller than a threshold of time interval ($T_{interval}$), the current loop closure edge would be added into the current cluster. Otherwise, a new cluster would be created. If the total time between the current loop closure edge ($t_{end}$) and the first loop closure edge ($t_{start}$) is greater than a threshold of total time ($T_{total}$), the current loop closure edge would be added into a new cluster. An explanation of clustering is shown in Fig.\ref{fig_clustering_loop_edges}. Cluster 1 is a normal situation where time interval and total time are both greater than threshold of time interval ($T_{interval}$) and total time ($T_{total}$) up to the next loop closure edge. Total time greater than threshold leads to divided consecutive loop closure edges into cluster 2 and cluster 3. Since time interval is greater than threshold, cluster 4 and cluster 5 are created.

\subsection{Scene Integration}
\begin{figure}[!ht]
\centering
\includegraphics[width=3.5in]{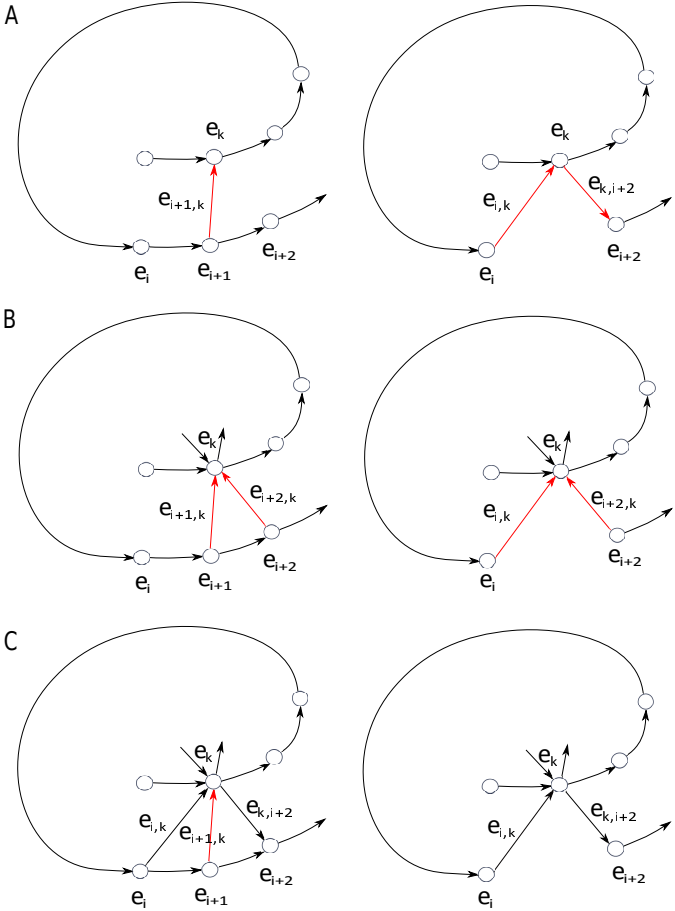}
\caption{Scene Integration. Standard cognitive map is on the left. Compact cognitive map is on the right. (A), (B) and (C) show three different cases we remove revisited vertices. Red arrows stand for the loop closure edges. }
\label{fig_remove_revisited_vertices}
\end{figure}

When mammals revisit familiar environments, the current scene would be integrated with the previous old scene stored in the memory.
For the robot, to achieve long-term mapping, another important thing is to control the size of the cognitive map bounded by the size of the explored environment and dependent on the exploration time. Alternatively, the size of the map does not grow unless the unknown environment is explored. Therefore, when the robot revisits familiar image views, redundant vertices should not be added and the same scenes are integrated.
Since the loss function is used in our formulation of non-linear least squares optimization problem, the influence of incorrect constraints would be reduced. After optimization with Ceres, outliers would not be correctly optimized like correct edges. That is why we remove revisited vertices after optimization, not avoid adding redundant vertices to begin with~\cite{johannsson2013temporally}. We will not focus on outliers problem, since robustness is another topic.
We illustrate this reduction approach in Fig.\ref{fig_remove_revisited_vertices}. Red arrows stand for loop closure edges. In Fig.\ref{fig_remove_revisited_vertices}A, we remove vertex $e_{i+1}$, and merge three edges $e_{i,i+1}$, $e_{i+1,i+2}$ and $e_{i+1,k}$ into two edges $e_{i,k}$ and $e_{k,i+2}$. Situation in Fig.\ref{fig_remove_revisited_vertices}B often happens, where multiple vertices connect to a same vertex. As for Fig.\ref{fig_remove_revisited_vertices}C, redundant vertices and edges are just removed.

\subsection{Removing Short Edges}
\begin{figure}[!ht]
\centering
\includegraphics[width=3in]{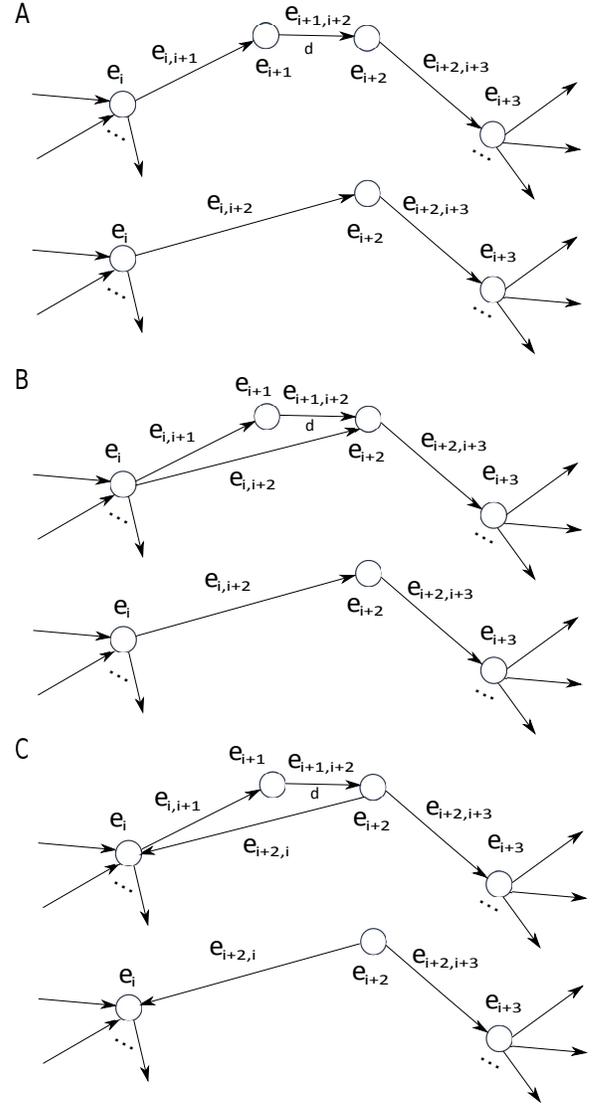}
\caption{Removing short edges. Standard cognitive map is on the top. Compact cognitive map is on the bottom. Edge $e_{i+1,i+2}$ length is smaller than threshold in (A). Vertex $e_{i+1}$ is removed. Edge $e_{i,i+1}$ and edge $e_{i+1,i+2}$ are merged into $e_{i,i+2}$. There exist multiple constraints between vertex $e_{i+1}$ and $e_{i+2}$ in (B) and (C). We remove edges $e_{i,i+1}$ and $e_{i+1,i+2}$, and vertex $e_{i+1}$.}
\label{fig_remove_short_edges}
\end{figure}

When the robot moves, due to motion noise and measuring error, redundant vertices might be created. After loop closures, extra constraints also need to be removed. These points represent same information. One of these points will be kept, and others are removed. Fig.\ref{fig_remove_short_edges} shows how we remove additional vertices and edges with threshold of edges length.

\subsection{Visual SLAM System}
Our compact cognitive mapping approach is demonstrated on a brain-inspired visual SLAM system. The presented visual SLAM system is improved from our previous work described in~\cite{taiping_Neurobiol_2017}. Here, we employ the above mentioned compact cognitive mapping technique to replace the graph relaxation algorithm in the experience map~\cite{ball_openratslam:_2013}.

\section{Experimental Results}
\label{results}
In this section, we demonstrate our compact cognitive mapping technique on a publicly available open-source dataset, iRat Australia dataset~\cite{ball_irat:_2010}. Intelligent Rat animat technology, iRat, is a small mobile robot which is a tool to study navigation and embodied cognition for robotic and neuroscience teams. iRat has a similar size and shape like a rodent. This iRat Australia dataset images are obtained by web camera. The iRat ROS bag provides camera images, odometry messages, and overhead images.

We compare our approaches with the performance of the method when no vertices and edges are discarded, which is referred to as the standard method and the corresponding cognitive map is referred to as the standard cognitive map. Compact cognitive mapping process to varying degree are shown in video s1, s2, s3 in the supplementary materials.

\subsection{Cognitive Map}
\begin{figure}[htbp]
\centering
\includegraphics[width=2.7in]{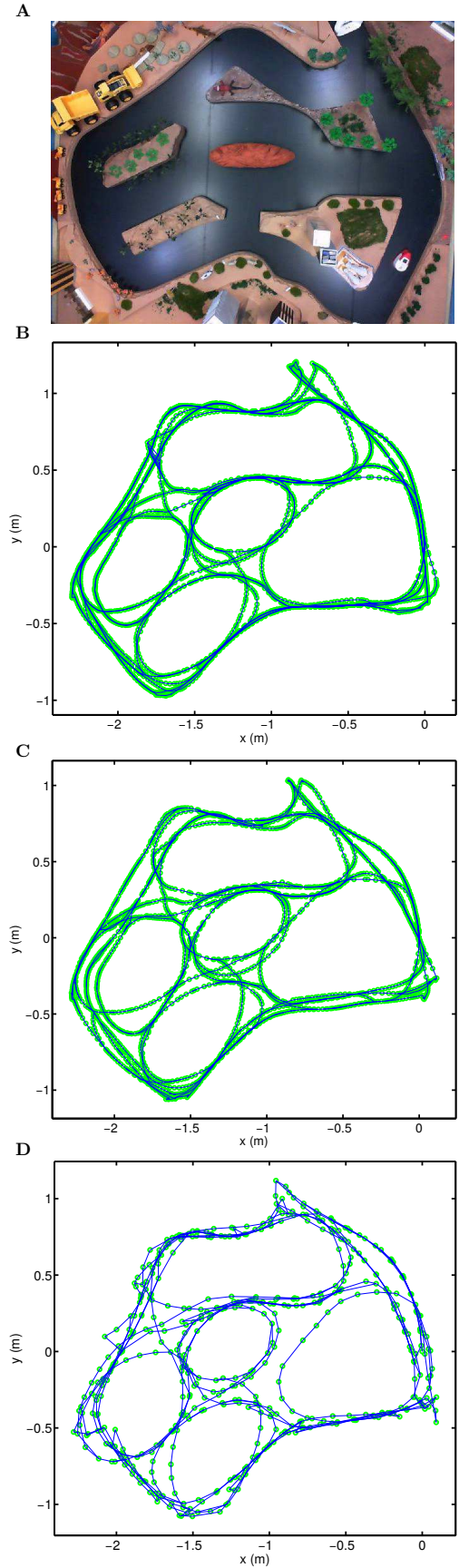}
\caption{Cognitive Map. Green dots are vertices of cognitive map. Blue thin lines are edges between connected vertices. (A) The overhead view of the explored environment. (B) shows the standard cognitive map with 3911 vertices and 5184 edges. (C) shows the compact cognitive map with 2191 vertices and 2152 edges by the step, scene integration. (D) shows the compact cognitive map with 497 vertices and 602 edges by steps: adding sparse vertices and edges, and scene integration.}
\label{fig_cognitive_map}
\end{figure}

To better show the ability of our approach to achieve vertex and edge sparsification, we mainly consider the influence of two steps: adding sparse sequential vertices and edges through neighborhood fields, and scene integration. Clustering loop closure edges is to reduce the number of global optimization execution. Batch global optimization with parallel computing based on OpenMP ensures that the optimization problem can be quickly solved. Since the frequency of batch global optimization execution is not high, there is enough time for batch global optimization, even for the cognitive map with large size. Therefore, clustering loop closure edges and batch global optimization would not decrease the size of the cognitive map, but for optimization efficiency. As for the step, removing short edges, it is mainly applied to reduce the possible noise.
In order to qualitatively compare the compact cognitive map with the real environment, the overhead of the explored environment is shown in~Fig.\ref{fig_cognitive_map}A. 

The standard cognitive map created by the visual SLAM system is shown in Fig.\ref{fig_cognitive_map}B. Green dots are vertices of cognitive map. Blue thin lines are edges between connected vertices.
There are 3911 vertices and 5184 edges included in the standard cognitive map. The standard mapping process is shown in video s1 in supplementary materials. The process of clustering loop closure edges and batch global optimization can be clearly seen from the video s1 in the supplementary materials.

When we add the step, scene integration, the total number of vertices in the compact cognitive map reduces from 3911 to 2191, edges from 5182 to 2152. The compact cognitive map is shown in Fig.\ref{fig_cognitive_map}C. Although number of vertices in Fig.\ref{fig_cognitive_map}C are reduced nearly twice compared with number of vertices in Fig.\ref{fig_cognitive_map}B, the standard cognitive map Fig.\ref{fig_cognitive_map}B is almost identical to the compact cognitive map in Fig.\ref{fig_cognitive_map}B compared by naked eyes. The mapping process corresponding to Fig.\ref{fig_cognitive_map}C is shown in video s2 in the supplementary materials.

\begin{figure}[htbp]
\begin{tabular}{l}
A\\
\begin{minipage}[b]{0.7\linewidth}
\includegraphics[width=3.3in]{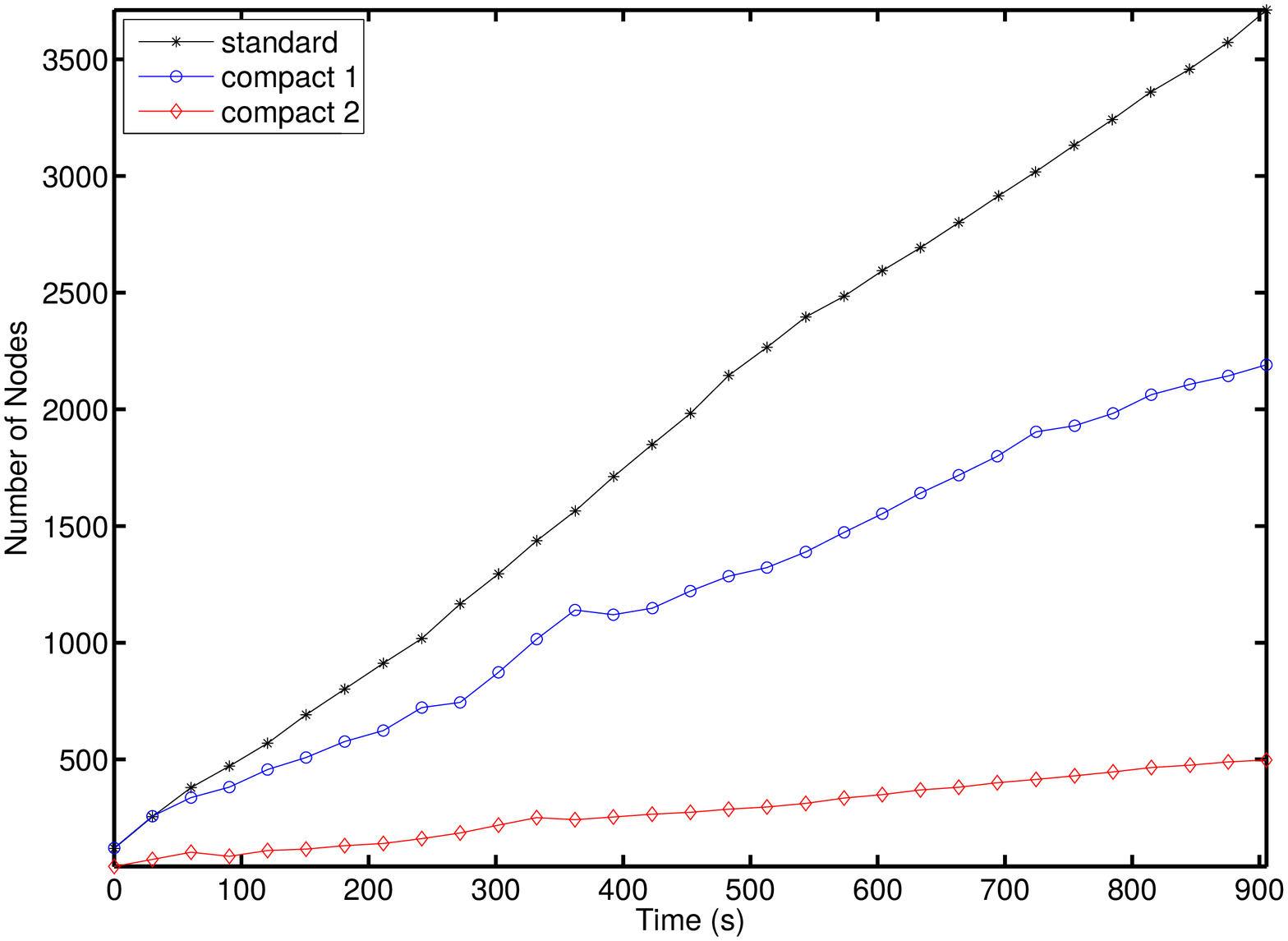}
\end{minipage}\\
\vspace{0.1cm}\\
B\\
\begin{minipage}[b]{0.7\linewidth}
\includegraphics[width=3.3in]{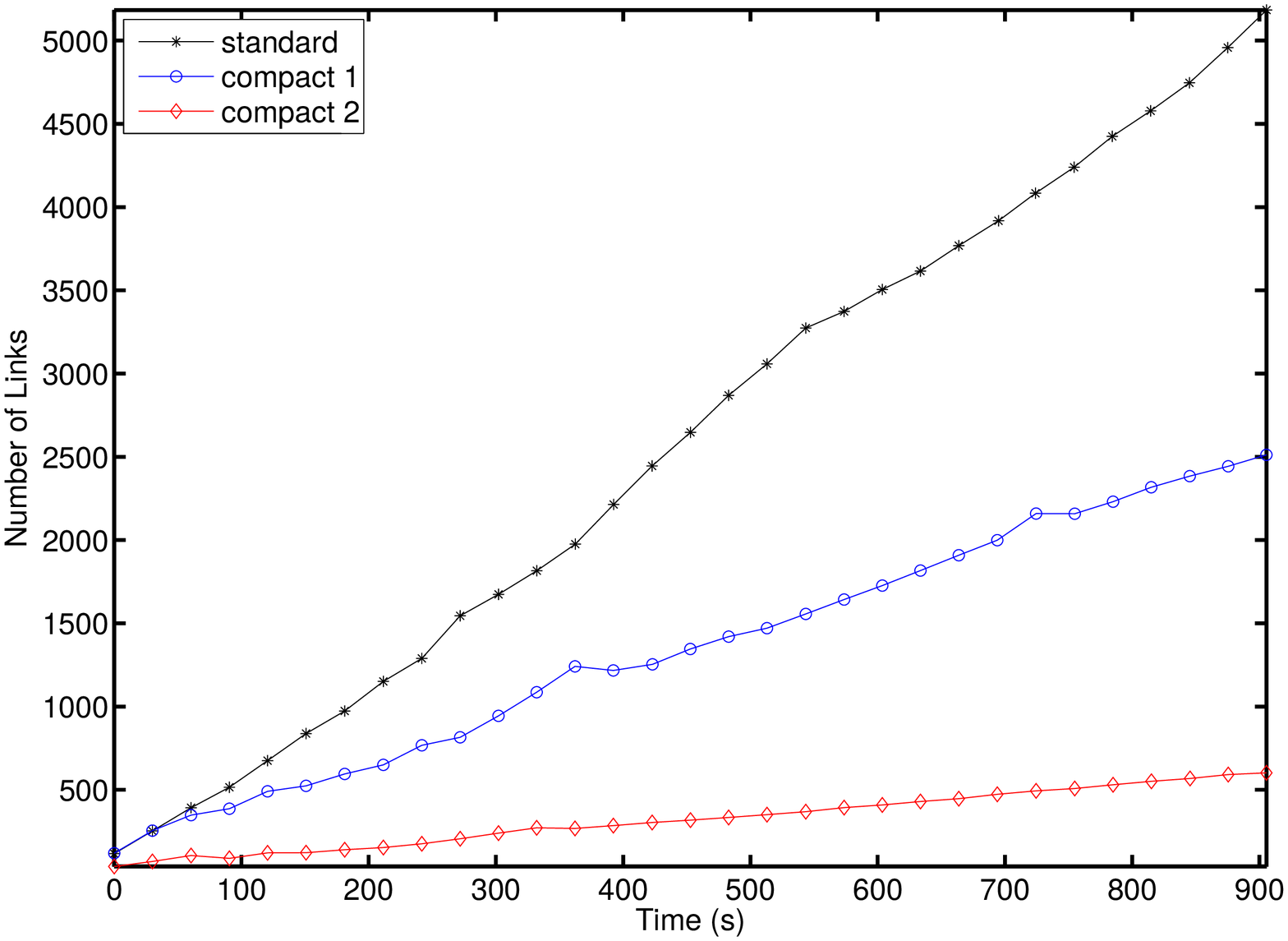}
\end{minipage}\\
\end{tabular}
\caption{Size of cognitive map when exploring the environment. Legend standard, compact 1, and compact 2 correspond to cognitive map in Fig.\ref{fig_cognitive_map}A, B and C, respectively. (A) shows number of vertices growing over time. (B) shows number of edges growing over time. }
\label{fig_size_of_map}
\end{figure}

Fig.\ref{fig_cognitive_map}D shows the compact cognitive map achieved by two steps, adding sparse sequential vertices and edges, and scene integration (see video s3 in supplementary materials). The number of vertices is reduced from 3911 to 497, edges from 5182 to 602, compared with the standard cognitive map. Here, neighborhood fields are employed to bound distance between two vertices and restrict the angle of rotation. The same weights for translation $\alpha = 10.0$ and rotation $\beta = 10.0$ are used to describe movement information. The threshold of neighborhood fields is set to $3.746$ through the debugging experience to determine whether adding a new vertex to the cognitive map.
The compact cognitive map in Fig.\ref{fig_cognitive_map}D correctly represents all loop closures and intersections, although the compact cognitive map in Fig.\ref{fig_cognitive_map}D is slight different from the cognitive map in Fig.\ref{fig_cognitive_map}B and C. The compact cognitive map built by our method is consistent with the standard cognitive map.

All in all, as you can see in Fig.\ref{fig_cognitive_map}, comparing with the overhead view of the explored environment in Fig.\ref{fig_cognitive_map}A, except little rotation offset, the cognitive map can successfully represent the overall layer of the explored environment, including the compact cognitive map in Fig.\ref{fig_cognitive_map}C and D.

\subsection{Size of Cognitive Map}
Depicted in Fig.\ref{fig_size_of_map}, number of vertices and edges in the cognitive map grows up as a function of exploration time. Standard cognitive map is the map without discarding vertices and edges. Dark line with star markers ($*$) in Fig.\ref{fig_size_of_map}A and B stands for the number of vertices and edges growing up to 3911 and 5182 in the standard cognitive map, respectively. After adding the step, scene integration, number of vertices and edges are reduced to 2191 and 2152 respectively, which is described by blue line with circular markers (o). Redline with diamond markers ($\Diamond$) shows the final compact cognitive map, whose size is controlled by two steps, adding sparse vertices and edges, and scene integration, reduced to 497 vertices and 602 edges.

\subsection{Clustering of Loop Closure edges}
As loop closure edges are created in the experience map, they are incrementally assigned to an existing cluster when they satisfy the requirement in Fig.\ref{fig_clustering_loop_edges}, or a new cluster. 
\begin{figure}[!ht]
\centering
\includegraphics[width=3.4in]{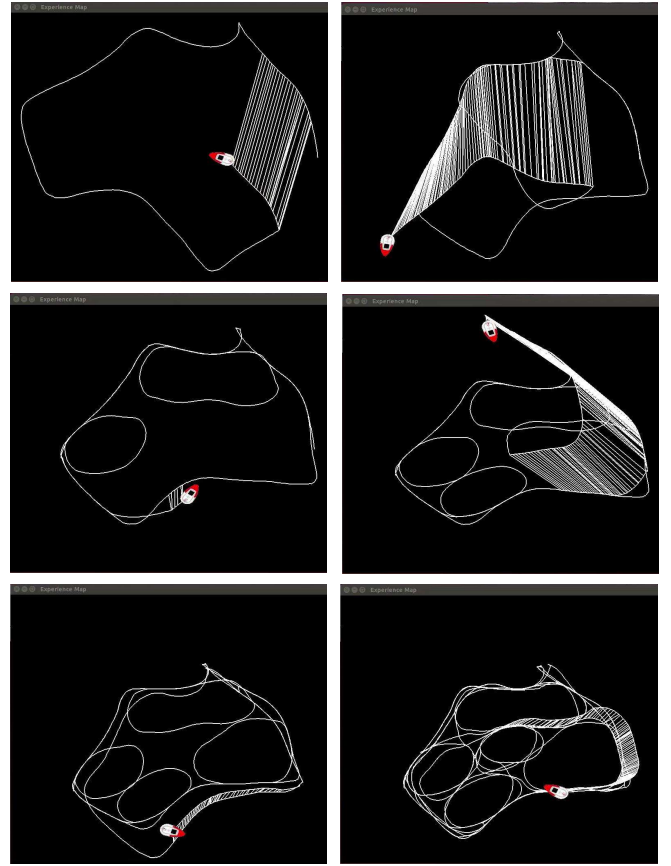}
\caption{Clustering of loop closure edges. Six loop closure clusters are selected throughout the mapping process, which includes different cluster size.}
\label{fig_screenshot_cluster_loop_closures}
\end{figure}
Every cluster grows dynamically, which allows loop closure edges with a similar trajectory to group together. In our experiment, for iRat Australia dataset, we consider the cluster total time interval threshold $T_{total}$ less than 100 seconds to be a part of the same cluster, and the loop cluster edge time interval threshold $T_{interval}$ greater than 2 seconds to be a part of a new cluster. Screenshots of clustering loop closures are shown in Fig.\ref{fig_screenshot_cluster_loop_closures}. Every cluster size is different from each other, which depends on the trajectory of the robot. For every cluster, only once batch global optimization is required to ensure the computational efficiency.

\section{Discussion}
\label{discussion}
In this study, we proposed a brain-inspired compact cognitive mapping solution to control the growth of the size of cognitive map. We implemented our solution in a monocular visual SLAM system, and demonstrated that our cognitive mapping system could successfully build a compact cognitive map and correctly represent the overall layer of the environment as the standard one from an open-source iRat Australia dataset (see video s1, s2, and s3 in the supplementary materials).

Our approach allows to gently trade off accuracy for computational cost. Neighborhood fields, which are inspired by neighborhood cells, can be employed to adjust the extent of cognitive map sparseness, which can be seen in Fig.\ref{fig_cognitive_map}B and C. Although in Fig.\ref{fig_cognitive_map}A, B and C (corresponding to video s1, s2, s3 in the supplementary materials), sizes of cognitive map differ greatly from each other, every cognitive map correctly represents the overall layer of the environment.

In our approach, we first formulate the global optimization of the cognitive map as a non-linear least squares problems. Compared with graph relaxation~\cite{ball_openratslam:_2013}, a fast sparse solver is applied to solve the non-linear least squares problems with high performance in Ceres solver.
Second, clustering loop closure edges is used to group loop closure edges with similar trajectory together. Only for every cluster, we use Ceres solver to compute a solution of global optimization, which greatly improves operational efficiency. Parallel processing with multicores based OpenMP boosts computing speed of global optimization.
Third, inspired from neurobiological experiments, a method based on neighborhood fields, adding sparse sequential vertices and edges, is proposed to perform compact cognitive mapping. As the neighborhood fields is determined by movement information including translation and rotation, along a straight road, the neighborhood field increases slowly means that less information is needed to remember, and whereas, when making a turn, especially crossroad, the neighborhood field increases rapidly means that more information should be required.
Finally, revisiting the same places would not be memorized many times, but integrated with the previous memory in the mammalian brain, which would not increase the size of the cognitive map, unless the unknown environment is explored. Scene integration is effective to maintain the size of cognitive map when the exploration environment is not altered or increased. 

Compared with the laser-based pose graph SLAM compressed by information-theory~\cite{ila2010information, kretzschmar2012information}, our approach is less theoretical, but pragmatic and efficient for for the cognitive map.
Temporally scalable stereo-vision-based SLAM by Johannsson and Leonard \cite{johannsson2013temporally} is also relevant to our approach. Avoid adding redundant vertices at the beginning, not marginalization, is applied to reduce pose graph, and then iSAM is used to perform map state estimation. In our approach, we cluster loop closure edges, after optimization, then revisited vertices are removed. And we formulate a nonlinear least square problem to the global optimization using a general fast non-linear least square solver, i.e. Ceres solver.
A brain-inspired monocular visual SLAM decreases the size of experience map by removing experiences to maintain one experience per grid square density by partitioning the environment~\cite{milford2010persistent}. We achieve sparsification of the cognitive map by introducing the concept of the neighborhood fields. The movement information with topographical orientation is applied to sparsify sequential vertices and edges in the cognitive map. Besides, we also cluster loop closures edges and perform batch optimization with parallel computing to ensure real-time performance. 

There are still several limitations that remain to be explored in our study. First, our algorithm needs to be tested longer time in the larger environment. Second, although the loss function reduces the influence of outliers, it can not entirely guarantee the quality of map estimation during the global optimization of the cognitive map.

In the future, we would demonstrate our compact cognitive mapping system on the larger environment in a sufficiently long time and deploy it to the robot. Also, a robustness method is considered to reject incorrect loop closure constraints.

\section{Conclusion}
\label{conclusion}
In short, we proposed a brain-inspired compact cognitive mapping system. Inspired from neurobiological experiments, the concept of neighborhood field and scene integration are applied to achieve sparsification of the cognitive map, and redundant vertices and edges are removed from our cognitive map. 
Furthermore, imitating the way that mammals control the size of the map is possible to develop practical algorithms to store the map during long-term operation for robot operation in complex, large scale, and dynamic environments.


\ifCLASSOPTIONcaptionsoff
  \newpage
\fi

\bibliographystyle{IEEEtran}
\bibliography{IEEEabrv,compact}

\begin{thebibliography}{10}
\providecommand{\url}[1]{#1}
\csname url@samestyle\endcsname
\providecommand{\newblock}{\relax}
\providecommand{\bibinfo}[2]{#2}
\providecommand{\BIBentrySTDinterwordspacing}{\spaceskip=0pt\relax}
\providecommand{\BIBentryALTinterwordstretchfactor}{4}
\providecommand{\BIBentryALTinterwordspacing}{\spaceskip=\fontdimen2\font plus
\BIBentryALTinterwordstretchfactor\fontdimen3\font minus
  \fontdimen4\font\relax}
\providecommand{\BIBforeignlanguage}[2]{{%
\expandafter\ifx\csname l@#1\endcsname\relax
\typeout{** WARNING: IEEEtran.bst: No hyphenation pattern has been}%
\typeout{** loaded for the language `#1'. Using the pattern for}%
\typeout{** the default language instead.}%
\else
\language=\csname l@#1\endcsname
\fi
#2}}
\providecommand{\BIBdecl}{\relax}
\BIBdecl

\bibitem{tolman_cognitive_1948}
E.~C. Tolman, ``Cognitive maps in rats and men.'' \emph{Psychological review},
  vol.~55, no.~4, p. 189, 1948.

\bibitem{zeng2017cognitive}
T.~Zeng and B.~Si, ``Cognitive mapping based on conjunctive representations of
  space and movement,'' \emph{Frontiers in Neurorobotics}, vol.~11, 2017.

\bibitem{kretzschmar2012information}
H.~Kretzschmar and C.~Stachniss, ``Information-theoretic compression of pose
  graphs for laser-based slam,'' \emph{The International Journal of Robotics
  Research}, vol.~31, no.~11, pp. 1219--1230, 2012.

\bibitem{cadena_past_2016}
C.~Cadena, L.~Carlone, H.~Carrillo, Y.~Latif, D.~Scaramuzza, J.~Neira, I.~Reid,
  and J.~J. Leonard, ``Past, present, and future of simultaneous localization
  and mapping: {Toward} the robust-perception age,'' \emph{IEEE Transactions on
  Robotics}, vol.~32, no.~6, pp. 1309--1332, 2016.

\bibitem{okeefe_hippocampus_1971}
J.~O'Keefe and J.~Dostrovsky, ``The hippocampus as a spatial map. {Preliminary}
  evidence from unit activity in the freely-moving rat,'' \emph{Brain
  research}, vol.~34, no.~1, pp. 171--175, 1971.

\bibitem{hafting_microstructure_2005}
T.~Hafting, M.~Fyhn, S.~Molden, M.-B. Moser, and E.~I. Moser, ``Microstructure
  of a spatial map in the entorhinal cortex,'' \emph{Nature}, vol. 436, no.
  7052, pp. 801--806, 2005.

\bibitem{taube_head-direction_1990}
J.~S. Taube, R.~U. Muller, and J.~B. Ranck, ``Head-direction cells recorded
  from the postsubiculum in freely moving rats. {I}. {Description} and
  quantitative analysis,'' \emph{The Journal of neuroscience}, vol.~10, no.~2,
  pp. 420--435, 1990.

\bibitem{kropff_speed_2015}
E.~Kropff, J.~E. Carmichael, M.-B. Moser, and E.~I. Moser, ``Speed cells in the
  medial entorhinal cortex,'' \emph{Nature}, 2015.

\bibitem{lever_boundary_2009}
C.~Lever, S.~Burton, A.~Jeewajee, J.~O'Keefe, and N.~Burgess, ``Boundary vector
  cells in the subiculum of the hippocampal formation,'' \emph{The journal of
  neuroscience}, vol.~29, no.~31, pp. 9771--9777, 2009.

\bibitem{mcnaughton_path_2006}
B.~L. McNaughton, F.~P. Battaglia, O.~Jensen, E.~I. Moser, and M.-B. Moser,
  ``Path integration and the neural basis of the'cognitive map','' \emph{Nature
  Reviews Neuroscience}, vol.~7, no.~8, pp. 663--678, 2006.

\bibitem{moser_place_2008}
E.~I. Moser, E.~Kropff, and M.-B. Moser, ``Place cells, grid cells, and the
  brain's spatial representation system,'' \emph{Annual review of
  neuroscience}, vol.~31, 2008.

\bibitem{moser_place_2015}
M.-B. Moser, D.~C. Rowland, and E.~I. Moser, ``Place cells, grid cells, and
  memory,'' \emph{Cold Spring Harbor perspectives in biology}, vol.~7, no.~2,
  p. a021808, 2015.

\bibitem{bos2017perirhinal}
J.~J. Bos, M.~Vinck, L.~A. van Mourik-Donga, J.~C. Jackson, M.~P. Witter, and
  C.~M. Pennartz, ``Perirhinal firing patterns are sustained across large
  spatial segments of the task environment,'' \emph{Nature Communications},
  vol.~8, 2017.

\bibitem{lu1997globally}
F.~Lu and E.~Milios, ``Globally consistent range scan alignment for environment
  mapping,'' \emph{Autonomous robots}, vol.~4, no.~4, pp. 333--349, 1997.

\bibitem{kummerle_g_2011}
R.~Kümmerle, G.~Grisetti, H.~Strasdat, K.~Konolige, and W.~Burgard, ``g 2 o:
  {A} general framework for graph optimization,'' in \emph{Robotics and
  {Automation} ({ICRA}), 2011 {IEEE} {International} {Conference} on}.\hskip
  1em plus 0.5em minus 0.4em\relax IEEE, 2011, pp. 3607--3613.

\bibitem{agarwal2012ceres}
S.~Agarwal, K.~Mierle \emph{et~al.}, ``Ceres solver,'' 2012.

\bibitem{ila2010information}
V.~Ila, J.~M. Porta, and J.~Andrade-Cetto, ``Information-based compact pose
  slam,'' \emph{IEEE Transactions on Robotics}, vol.~26, no.~1, pp. 78--93,
  2010.

\bibitem{carlevaris2013generic}
N.~Carlevaris-Bianco and R.~M. Eustice, ``Generic factor-based node
  marginalization and edge sparsification for pose-graph slam,'' in
  \emph{Robotics and Automation (ICRA), 2013 IEEE International Conference
  on}.\hskip 1em plus 0.5em minus 0.4em\relax IEEE, 2013, pp. 5748--5755.

\bibitem{mazuran2016nonlinear}
M.~Mazuran, W.~Burgard, and G.~D. Tipaldi, ``Nonlinear factor recovery for
  long-term slam,'' \emph{The International Journal of Robotics Research},
  vol.~35, no. 1-3, pp. 50--72, 2016.

\bibitem{johannsson2013temporally}
H.~Johannsson, M.~Kaess, M.~Fallon, and J.~J. Leonard, ``Temporally scalable
  visual slam using a reduced pose graph,'' in \emph{Robotics and Automation
  (ICRA), 2013 IEEE International Conference on}.\hskip 1em plus 0.5em minus
  0.4em\relax IEEE, 2013, pp. 54--61.

\bibitem{milford2010persistent}
M.~Milford and G.~Wyeth, ``Persistent navigation and mapping using a
  biologically inspired slam system,'' \emph{The International Journal of
  Robotics Research}, vol.~29, no.~9, pp. 1131--1153, 2010.

\bibitem{ball_openratslam:_2013}
D.~Ball, S.~Heath, J.~Wiles, G.~Wyeth, P.~Corke, and M.~Milford,
  ``{OpenRatSLAM}: an open source brain-based {SLAM} system,'' \emph{Autonomous
  Robots}, vol.~34, no.~3, pp. 149--176, 2013.

\bibitem{taiping_Neurobiol_2017}
T.~Zeng and B.~Si, ``Neurobayesslam: Neurobiologically inspired bayesian
  integration of multisensory information for robot navigation,''
  \emph{manuscript}, 2017.

\bibitem{ball_irat:_2010}
D.~Ball, S.~Heath, G.~Wyeth, and J.~Wiles, ``{IRat}: {Intelligent} rat animat
  technology,'' in \emph{Proceedings of the 2010 {Australasian} {Conference} on
  {Robotics} and {Automation}}, 2010, pp. 1--3.

\end{thebibliography}





\vfill



\end{document}